\documentclass[conference]{ifacconf}
\usepackage{resizegather}
\usepackage{graphicx}
\usepackage{natbib}
\usepackage{graphics} 
\usepackage{epsfig} 
\usepackage{epstopdf}
\usepackage[abs]{overpic}
\usepackage{setspace}
\usepackage{mathtools}
\usepackage{algorithmicx,algpseudocode}
\usepackage{amsmath,amsfonts,amssymb}
\newcommand{\argmin}{\operatornamewithlimits{argmin}}
\usepackage{accents}
\usepackage{tikz}
\usepackage{theorem}

\newcounter{algno} 
\mathtoolsset{showonlyrefs}
\newcounter{asno} 
\theoremstyle{definition}

\newcounter{rem} 
\begin{document}
\begin{frontmatter}
\title{Distributed Path Planning for Executing Cooperative Tasks with Time Windows}

\author[First]{Raghavendra Bhat}, 
\author[First]{Yasin Yaz{\i}c{\i}o\u{g}lu}, and 
\author[Second]{Derya Aksaray}

\address[First]{Department of Electrical and Computer Engineering, University of Minnesota,
Minneapolis, Minnesota 55455.
Email: \tt bhatx075@umn.edu, ayasin@umn.edu.}
\address[Second]{Department of Aerospace Engineering and Mechanics, University of Minnesota, Minneapolis, Minnesota 55455.  Email: \tt daksaray@umn.edu }

\begin{abstract}                 
We investigate the distributed planning of robot trajectories for optimal execution of cooperative tasks with time windows. In this setting, each task has a value and is completed if sufficiently many robots are simultaneously present at the necessary location within the specified time window. Tasks keep arriving periodically over cycles. The task specifications (required number of robots, location, time window, and value) are unknown a priori and the robots try to maximize the value of completed tasks by planning their own trajectories for the upcoming cycle based on their past observations in a distributed manner. Considering the recharging and maintenance needs, robots are required to start and end each cycle at their assigned stations located in the environment. We map this problem to a game theoretic formulation and maximize the collective performance through distributed learning. Some simulation results are also provided to demonstrate the performance of the proposed approach.
\end{abstract}

\begin{keyword}
Distributed control,  multi-robot systems, planning, game theory, learning
\end{keyword}

\end{frontmatter}





\section{Introduction}
 
\vspace{-4mm}
Teams of mobile robots provide  efficient and robust solutions in multitude of applications such as precision agriculture, environmental monitoring, surveillance, search and rescue, and warehouse automation. Such applications typically require the robots to complete a variety of spatio-temporally distributed tasks, some of which (e.g., lifting heavy objects) may require the cooperation of multiple robots. In such a setting, successful completion of the tasks require the presence of sufficiently many robots at the right location and time. 
Accordingly, the overall performance depends on a coordinated plan of robot trajectories. 

Planning is one of the fundamental topics in robotics (e.g., see \cite{Lavalle06, Elbanhawi14} and the references therein). For multi-robot systems, the inherent complexity arising from the exponential growth of the joint planning space usually renders the exact centralized solutions intractable. Unlike the standard formulations such as minimizing the travel time, the energy consumed, or the distance traveled while avoiding collisions, there is limited literature on planning trajectories for serving spatio-temporally distributed cooperative tasks. In \cite{Thakur13}, the authors investigate a problem where a team of robots allocate a given set of waypoints among themselves and plan paths through those
waypoints while avoiding obstacles and reaching their goal regions by specific deadliness. In \cite{Bhattacharya10}, the authors consider the distributed path planning for robots with preassigned tasks that can be served in any order or time.

In this work, we study a \emph{distributed  task execution} (DTE) problem, where a homogeneous team of mobile robots optimize their trajectories to maximize the total value of completed cooperative tasks that arrive periodically over time at different locations in a discretized environment. In this setting, each task is defined by the following specifications:  required number of robots, location, time window (arrival and departure times), and value. Tasks are completed if sufficiently many robots simultaneously spend one time-step at the necessary location within the corresponding time window. Tasks keep arriving periodically over cycles and their  specifications are unknown a priori. Robots are required to start and end each cycle at their assigned stations in the environment and they try to maximize the value of completed tasks by each of them planning its own trajectory  based on the observations from previous cycles. 

In order to tackle the challenges due to the possibly large scale  of the system (number of tasks and robots) and initially unknown task specifications, we investigate how the DTE problem can be solved via distributed learning. Such distributed coordination problems are usually solved using methods based on machine learning, optimization, and game theory (e.g., see \cite{Bu08, Boyd11,Marden09} and the references therein). Game theoretic methods have been used to solve various problems such as vehicle-target assignment (e.g., \cite{Marden09}), coverage optimization (e.g., \cite{Zhu13,Yazicioglu13NECSYS, Yazicioglu17TCNS}), or dynamic vehicle routing (e.g., \cite{Arsie09}). In this paper, we propose a game theoretic solution to the DTE problem by designing a corresponding game and utilizing a learning algorithm that drives the robots to configurations that maximize the global objective, i.e., the total value of completed tasks. In the proposed game, the action of each robot is defined as its trajectory during one cycle. We show that some feasible trajectories can never contribute to the global objective in this setting, no matter what the task specifications are. By excluding such trajectories, we obtain a game with a significantly smaller action space, which still contains the globally optimal combinations of trajectories but also facilitates faster learning due to its smaller size. Using the proposed method, robots spend an arbitrarily high percentage of cycles at the optimal combinations of trajectories in the long run. 

This paper is organized as follows: Section \ref{prob} presents the formulation of the distributed task execution problem. Section \ref{method} is on the proposed game theoretic formulation and solution. Some simulation results are presented in Section \ref{sims}. Finally, Section \ref{conc} concludes the paper.

\section{Problem Formulation}
\label{prob}
In this section, we introduce the distributed task execution (DTE) problem, where the goal is to have a homogeneous team of $n$ mobile robots, $R=\{r_1, r_2, \hdots, r_n\}$, optimize their trajectories to maximize the total value of completed cooperative tasks that arrive periodically and remain available only over specific time windows. 

We consider a discretized environment represented as a 2D grid, $P=\{1,2, \hdots , \bar{x}\} \times \{1,2, \hdots , \bar{y}\}$, where $\bar{x},\bar{y} \in \mathbb{N}$ denote the number of cells along the corresponding directions. In this environment, some of the cells may be occupied by obstacles, $P_O \subset P$, and the robots are free to move over the feasible cells ${P_F= P\setminus P_O}$. Within the feasible space, we consider $m$ stations, ${S=\{s_1, \hdots, s_m\} \subseteq  P_F}$, where the robots start and end in each cycle. Stations denote the locations where the robots recharge, go through maintenance when needed, and prepare for the next cycle. Each robot is assigned to a specific station (multiple can be assigned to the same station) and must return there by the end of each cycle. We assume that each cell represents a sufficiently large amount of space, hence any number of robots can be present in the same cell at the same time.

Each cycle consist of $T$ time steps and the trajectory of each robot $r_i \in R$  over a cycle is denoted as ${\bold{p}_i=\{p_i^0, p_i^1, \hdots, p_i^T \}}$. The robots can maintain their current position or move to any of the feasible neighboring cells within one time step. Specifically, if a robot is at some cell $p=(x,y)\in P_F$, at the next time step it has to be within $p$'s neighborhood on the grid $N(p)\subseteq P_F$, i.e.,
\begin{equation}
    \label{motion}
    N(p)= \{(x',y') \in P_F \mid |x'-x| \leq 1, |y'-y| \leq 1\}.
\end{equation}

\begin{figure}[htb]
\label{env}
\centering
\includegraphics[trim =0mm 0mm 0mm 0mm, clip,scale=0.5]{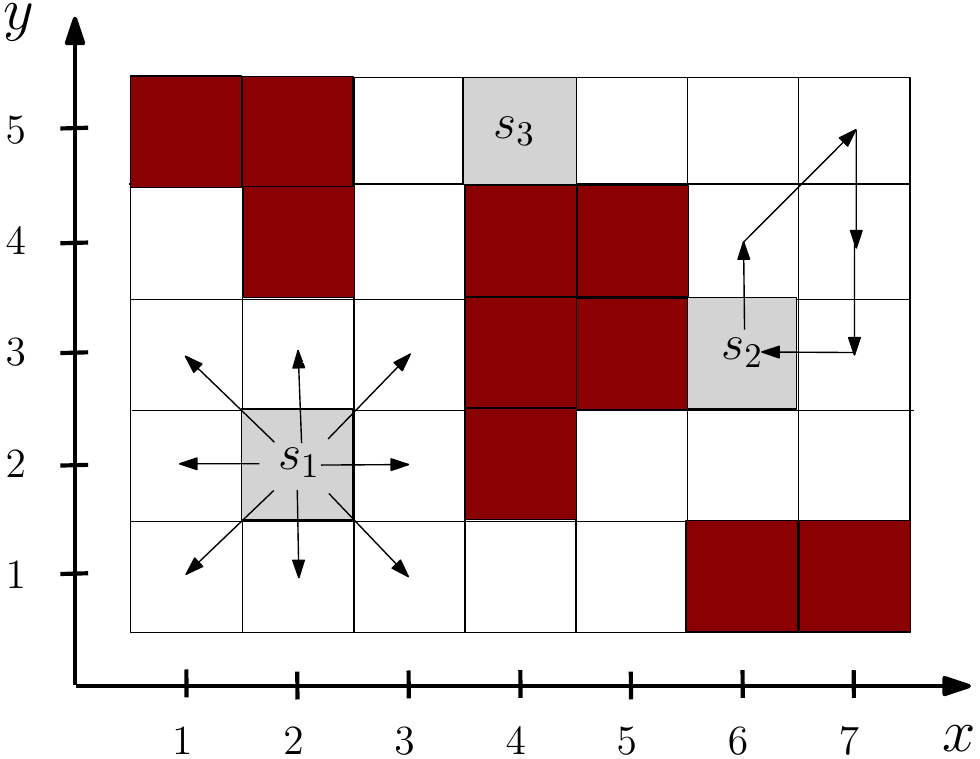}
\caption{\small A discretized environment with obstacles (red) and three stations (gray) is shown. Robots can move to any neighboring cell (non-obstacle) in one time step. On the left, the neighboring cells of station $s_1$ are shown via arrows. On the right, a feasible trajectory of length five is shown for a robot at station $s_2$. }
\end{figure}

For any robot $r_i\in R$, the set of feasible trajectories, $\bold{P}_i$, is defined as
\begin{equation}
    \label{traj-set}
    \bold{P}_i= \{\bold{p}_i \mid p_i^0=p_i^T=\sigma_i, p_i^{t+1} \in N(p_i^t), \;  t= 0, \hdots, T-1 \},
\end{equation}
where $\sigma_i \in S$ denotes the station of robot $r_i$. Accordingly, we use $\bold{P}=\bold{P}_1 \times \hdots \times \bold{P}_m$ to denote the combined set of feasible trajectories.
An example of an environment with some obstacles and three stations is illustrated in Fig. \ref{env} along with the examples of feasible motions and trajectories of robots.

Given such an environment and a set  of mobile robots, we consider a set of $k$ tasks $\tau=\{\tau_1, \tau_2, \hdots, \tau_k\}$, each of which is defined as tuple, $\tau_i= \{c_i^*, l_i, t^a_i,t^d_i,v_i\}$,
where $c_i^* \in \mathbb{N}$ is the required number of robots, $l_i \in P_F$ is the location, ${t^a_i<t^d_i \in \{0,\hdots,T\}}$ are the arrival and departure times,  and $v_i \in \mathbb{R}_+$ is the value. Accordingly, the task is completed if at least $c_i^*$ robots simultaneously spend one time step at $l_i$ within the time window $[t^a_i,t^d_i]$. More specifically, given the trajectories of all robots, $\bold{p}$, the set of completed tasks $\tau^* (\bold{p}) \subseteq \tau$ is defined as
 \begin{equation}
\label{taustar}
\tau^*(\bold{p}) = \{\tau_i \in \tau \mid \exists t \in [t^a_i,t^d_i-1], \; c_i(\bold{p},t) \geq c_i^* \},
\end{equation}
where $c_i(\bold{p},t) \in \{0, \hdots,n\}$ is the counter denoting the number of robots that stayed at $l_i\in P_F$ from time $t$ to $t+1$ in that cycle, i.e., 
\begin{equation}
\label{rtq}
c_i(\bold{p},t) =|\{r_i \in R \mid p_i^t = p_i^{t+1} = l_i\}|.
\end{equation}

Accordingly, each task can be completed within one time step if sufficiently many robots simultaneously stay at the corresponding location within the specified time window. This model captures a variety of tasks with time windows. Examples include placing a box in a shelf, where the weight of the box determines the number of robots needed, or aerial monitoring, where the required ground resolution determines the number of needed drones (higher resolution implies closer inspection, hence more drones needed to cover the area). As per this model, when a team of robots encounter a task, they are assumed to know how to achieve a proper low-level coordination (e.g., how multiple robots should move a box) so that the task is completed if there are sufficiently many robots.  Our main focus is on the problem of having sufficiently many robots in the right place and time window. In this regard, we quantify the performance resulting from the trajectories of all robots, $\bold{p}$, as the total value of completed tasks, i.e.,
\begin{equation}
\label{obj}
f(\bold{p})= \sum_{\tau_i \in \tau^*}v_i.
\end{equation}

We consider a setting where the tasks are unknown a priori. Accordingly, the robots are expected to improve the overall performance by updating their trajectories over the cycles based on their observations. In such a setting, $\bold{p}(t)$ denotes the trajectories at the $t^{th}$ cycle for $t\in \{0,1,\hdots\}$, and we are interested in the resulting long-run average performance. Accordingly, for any infinite sequence of robot trajectories over time (cycles), we quantify the resulting performance as
\begin{equation}
\label{obj-t}
\liminf_{t^* \to \infty} \frac{1}{t^*+1} \sum_{t=0}^{t^*}f(\bold{p}(t)).
\end{equation}

We are interested in optimizing the long-run average performance via a distributed learning approach, where each robot $r_i \in R$ independently plans its own trajectory for the upcoming cycle  based on its past observations. 


\section{Proposed Method}
\label{method}
We will present a game theoretic solution to the DTE problem. We first provide some game theory preliminaries.

\subsection{Game Theory Preliminaries}
A finite \emph{strategic game} $\Gamma= (I, A, U)$ has three components: (1) a set of \emph{players (agents)} $I=\{1, 2, \hdots, m\}$, (2) an \emph{action space} $A= A_1 \times A_2 \times \hdots\ \times A_m$, where each $A_i$ is the \emph{action set} of player $i$, and (3) a set of \emph{utility functions} $U= \{U_1, U_2, \hdots , U_m\}$, where each $U_i : A \mapsto \Re$ is a mapping from the action space to real numbers.   

For any \emph{action profile} $a \in A$, we use $a_{-i}$ to denote the actions of players other than $i$. Using this notation, an action profile $a$ can also be represented as $a=(a_i,a_{-i})$.
An action profile $a^* \in A$ is called a \emph{Nash equilibrium} if
\begin{equation}
\label{NE}
 U_i(a_i^*, a_{-i}^*) = \max_{a_i \in A_i} U_i(a_i,a_{-i}^*), \; \forall i \in I.
\end{equation}


A class of games that is widely utilized in cooperative control problems is the \emph{potential games}. A game is called a potential game if there exists a \emph{potential function}, $\phi : A \mapsto \Re$, such that the change of a player's utility resulting form its unilateral deviation from an action profile equals the resulting change in $\phi$. More precisely, for each player $i$, for every $a_i, a'_i \in A_i$, and for all $a_{-i} \in A_{-i}$,  
\begin{equation}
\label{PG}
 U_i(a'_i, a_{-i})- U_i(a_i, a_{-i}) = \phi(a'_i, a_{-i})- \phi(a_i, a_{-i}).
\end{equation}
When a cooperative control problem is mapped to a potential game, the game is designed such that its potential function captures the global objective (e.g., \cite{Marden09}). Such a design achieves some alignment between the global objective and the utility of each agent.

In game theoretic learning, starting from an arbitrary initial configuration, the agents repetitively play a game. At each step $t \in \{0, 1, 2, \hdots \}$, each agent $i \in  I$ plays an action $a_i(t)$ and receives some utility $U_i(a(t))$. 
In this setting, the agents update their actions in accordance with some learning algorithm. For potential games, there are many learning algorithms that provide convergence to a Nash equilibrium (e.g., \cite{Arslan07,Marden09} and  the references therein).  While any potential maximizer is a Nash equilibrium, a potential game may also have some suboptimal Nash equilibria. In that case, a learning algorithm such as \emph{log-linear learning} (LLL) presented in \cite{Blume93} can be used to drive the agents to the equilibria that maximize the potential function $\phi(a)$. Essentially, LLL is a noisy best-response algorithm, and it induces a Markov chain over the action space with a unique limiting distribution, $\mu^*_\epsilon$,  where $\epsilon$ denotes the noise parameter. When all agents follow LLL in potential games, the potential maximizers are the \emph{stochastically stable} states as shown in \cite{Blume93}. In other words, as the noise parameter $\epsilon$ goes down to zero, the limiting distribution, $\mu^*_\epsilon$, has an arbitrarily large part of its mass accumulated over the set of potential maximizers, i.e., 
\begin{equation}
\label{sstab}
\lim_ {\epsilon \rightarrow 0^+} \mu^*_{\epsilon}(a) >0 \Longleftrightarrow \phi(a) \geq \phi(a'), \forall a' \in A.
\end{equation}

To execute LLL, only one agent should  update its action at each step. Furthermore, each agent should be able to compute its current utility as well as the hypothetical utilities it may gather by unilaterally switching to any other action. Alternatively, the payoff-based implementation of LLL presented in \cite{Marden12} yields the same limiting behavior without requiring single-agent updates and the computation of hypothetical utilities.

\vspace{-1mm}
\subsection{Game Design}
\vspace{-1mm}
In light of \eqref{sstab}, if the DTE problem is mapped to a potential game such that the potential function is equal to \eqref{obj}, then the long-run average performance given in \eqref{obj-t} can be made arbitrarily close to its maximum possible value by having robots follow  some learning algorithm such as LLL. In this section, we build such a game theoretic solution. We first design a corresponding game $\Gamma_{\text{DTE}}$ by defining the action space and the utility functions.

Since the impact of each agent on the overall performance is solely determined by its trajectory, it is rather expected to define the action set of each agent as some subset of the feasible trajectories, i.e., $A_i \subseteq \bold{P}_i$. One possible choice is setting $A_i= \bold{P}_i$, which allows the robots to take any feasible trajectory. However, it should be noted that the learning will involve robots searching within their feasible actions. Accordingly, defining a larger action space is likely to result in a slower learning rate, which is an important aspect in practice. Due to this practical concern, we will design a more compact action space which can yield the same long run  performance as the case with all feasible trajectories, yet approaches the limiting behavior much faster. The impact of this reduction in the size of action space will later be demonstrated through numerical simulations in Section \ref{sims}. 

\subsubsection{Action Space Design} While the set of possible trajectories, $\bold{P}_i$, grows exponentially with the cycle length $T$, a large number of those trajectories actually can never be a useful choice in the proposed setting, regardless of the task specifications and the trajectories of other robots. In particular, if a trajectory doesn't involve staying anywhere, i.e., there is no $t \in \{0, \hdots,T-1\}$ such that $p_i^{t}=p_i^{t+1}$, it is guaranteed that the robot $i$ will not be contributing to the global score in \eqref{obj} since it will not be helping with any task (no contribution to any counters in \eqref{rtq}). Furthermore, there are also many trajectories, for which there exist some other trajectory guaranteed to be equally useful or better for the global performance no matter what the task specifications or the trajectories of other robots are. In particular, if a trajectory $\bold{p}_i$ has all the stays some other trajectory $\bold{q}_i$ has, then removing $\bold{q}_i$ from the feasible options would not degrade the overall performance. By removing such inferior trajectories from the available options, we define a significantly smaller action set without any reduction in the achievable long-run performance:
\begin{align}\label{actset}
    A_i = \argmin_{A_i \in 2^{\bold{P}_i}} & \quad |A_i|\\
    s.t. & \quad \eqref{actionsetc1}, \eqref{actionsetc2}, \nonumber
\end{align}
where the constraints are
\begin{equation}
\label{actionsetc1}
\forall \bold{p}_i \in A_i, \exists t\in\{0,1,\hdots,T-1\}: p_i^{t}=p_i^{t+1},
\end{equation}
\begin{equation}
\label{actionsetc2}
\forall \bold{q}_i \in \bold{P}_i \setminus A_i, \exists \bold{p}_i \in A_i: p_i^{t}=p_i^{t+1}=q_i^{t}, \forall t: q_i^{t}=q_i^{t+1}.
\end{equation}

Accordingly, the action set of each robot $A_i$ is the smallest subset of its all feasible trajectories $\bold{P}_i$ such that 1) each trajectory $\bold{p}_i \in A_i$ should have at least one stay, i.e.,\eqref{actionsetc1}, and 2) for every excluded trajectory $\bold{q}_i \in \bold{P}_i \setminus A_i$, there should be a trajectory $\bold{p}_i \in A_i$ such that any stay in $\bold{q}_i$ is also included in $\bold{p}_i$, i.e., \eqref{actionsetc2}. It is worth emphasizing that this reduced action space maintains all the global optima.

\begin{lem}
\label{maxpreserved}
For any set of feasible trajectories $\bold{P}_i$ as in \eqref{traj-set} and the action sets as in \eqref{actset},
\begin{equation}
\label{lem11}
\max_{\bold{p} \in \bold{P}}f(\bold{p}) = \max_{\bold{p} \in A} f(\bold{p}).
\end{equation}
\end{lem}
\begin{pf}
Let $\bold{q} \in \bold{P} \setminus A$ be a maximizer of $f(\bold{p})$, i.e.,
\begin{equation}
\label{lem12}
     f(\bold{q}) = \max_{\bold{p} \in \bold{P}} f(\bold{p}).
\end{equation}
In light of \eqref{actionsetc2}, there exist $\bold{p} \in A$ such that, for every robot $i$, all the stays under $\bold{q}_i$ are also included in $\bold{p}_i$, i.e., for any $t \in\{0, \hdots, T-1\}$
\begin{equation}
\label{lem13}
   q_i^t= q_i^{t+1} \Rightarrow p_i^t= p_i^{t+1}=q_i^t.
\end{equation}
Accordingly, in light of \eqref{rtq} and \eqref{taustar}, all the tasks completed under $\bold{q}$ should be completed under $\bold{p}$ as well, i.e., 
\begin{equation}
\label{lem13}
\tau^*(\bold{q}) \subseteq \tau^*(\bold{p}).
\end{equation}
Hence, $f(\bold{p}) \geq f(\bold{q})$. Since $A \subseteq \bold{P}$, $f(\bold{p}) \geq f(\bold{q})$ and \eqref{lem12} together imply \eqref{lem11}.
\end{pf}


\emph{Example:} Consider the environment in Fig \ref{env}, and let each cycle consist of three time steps ($T=3$). In that case, the feasible trajectories for a robot $i$ stationed at $s_1$ in cell $(2,2)$ are all the cyclic routes of three hops ${\bold{p}_i=\{p_i^0, p_i^1,p_i^2,p_i^3\}}$ such that ${p_i^0=p_i^3=(2,2)}$. It can be shown that there are 49 such feasible trajectories. However, by solving \eqref{actset}, one can obtain an action set $A_i$ consisting  of only the following nine trajectories:

${\{(2,2),(1,1), (1,1), (2,2)\}}$,
${\{(2,2),(1,2), (1,2), (2,2)\}}$,\newline ${\{(2,2),(1,3), (1,3), (2,2)\}}$,   
${\{(2,2),(2,1), (2,1), (2,2)\}}$,
\newline
${\{(2,2),(2,3), (2,3), (2,2)\}}$, 
${\{(2,2),(3,1), (3,1), (2,2)\}}$,  \newline
${\{(2,2),(3,2), (3,2), (2,2)\}}$,
${\{(2,2),(3,3), (3,3), (2,2)\}}$, \newline
${\{(2,2),(2,2), (2,2), (2,2)\}}$.

Note that these are the trajectories that 1) move to one of the eight adjacent cells, stay there for one time step, and return to the station, or 2) stay at cell $(2,2)$ throughout the cycle,  which are indeed the only choices  that may result in helping with the completion of some task in this example. 

\subsubsection{Utility Design} We design the utility functions so that the total value of completed tasks becomes the potential function of the resulting game. To this end we employ the notion of \emph{wonderful life utility} presented in \cite{Tumer04}. Accordingly, we set the utility of each robot $i$ to the total value of completed tasks that would not have been completed if robot $i$ was removed from the system, i.e.,
\begin{equation}
\label{util}
U_i (\bold{p}) =  \sum_{\tau_j \in (\tau^*(\bold{p})\setminus \tau^*(\bold{p}_{-i})) }v_j,
\end{equation}
where $\tau^*(\bold{p})$, as defined in \eqref{taustar}, is the set of tasks completed given the trajectories of all robots and $\tau^*(\bold{p}_{-i})$ is the set of tasks completed when robot $i$ is excluded from the system. 

\begin{lem}
\label{potgam}
For any set of robots $R$ with the action space $A$ as per \eqref{actset}, the utilities in \eqref{util} lead to a potential game $\Gamma_{\emph{DTE}}= (R, A, U)$ with the potential function ${\phi(\bold{p})= f(\bold{p})}$,
where $f(\bold{p})$ is the total value of completed tasks as given in \eqref{obj}.
\end{lem}

\begin{pf}
Let $\bold{p}_i \neq \bold{p}_i' \in A_i$ be two possible trajectories for robot $i$, and let $\bold{p}_{-i}$ denote the trajectories of all other robots. Since removing a robot from the system cannot increase the number of robots present at each cell during the cycle, for each task $\tau_j$ we have $c_j (\bold{p},t) \geq c_j(\bold{p}_{-i},t)$ for the counters as defined in \eqref{rtq}. Accordingly, by removing robot $i$ from the system, the set of completed tasks can only shrink, i.e., $\tau^*(\bold{p}_{-i}) \subseteq \tau^*(\bold{p})$. Hence, the utility in \eqref{util} can be expressed as
 \begin{equation}
\label{potgam1}
U(\bold{p}_i,\bold{p}_{-i})=  \sum_{\tau_j \in \tau^*(\bold{p}_i,\bold{p}_{-i}) }v_j - \sum_{\tau_j \in  \tau^*(\bold{p}_{-i}) }v_j.
\end{equation}
Accordingly,
 \begin{eqnarray}
\label{potgam2}
U(\bold{p}_i,\bold{p}_{-i})-U(\bold{p}_i',\bold{p}_{-i}) &=&  \sum_{\tau_j \in \tau^*(\bold{p}_i,\bold{p}_{-i}) }v_j - \sum_{\tau_j \in \tau^*(\bold{p}_i',\bold{p}_{-i}) }v_j, \\
&=& f(\bold{p}_i,\bold{p}_{-i})-f(\bold{p}_i',\bold{p}_{-i}).
\end{eqnarray}
Consequently, $\Gamma_{\emph{DTE}}= (R, A, U)$ is a potential game with the potential function $\phi(\bold{p})= f(\bold{p})$.
\end{pf}

Note that the utility in \eqref{util} can be computed by each robot based on local information. For every completed task $\tau_j \in \tau^* (\bold{p})$, let  $t^*_j(\bold{p})$ denote the time when the participating robots started execution, i.e.,
 \begin{equation}
     \label{compt}
     t^*_j(\bold{p}) = min \{t \in [t^a_j, t^d_j-1] \mid c_j (\bold{p},t) \geq c^*_j\}.
 \end{equation}
 
Each robot involved in the execution of $\tau_j$ needs to know the value $v_j$ and whether the task could have been completed without itself to compute the amount of reward it receives from this task (0 or $v_j$). To assess whether the task could have been completed without itself, the robot needs to know the required number of robots $c_j^*$, and the number of robots that stay at $l_j$ starting from $t^*_j$ until the last point the task can be completed, i.e., $c_j (\bold{p},t)$ for all $t\in \{t^*_j, \hdots, t^d_j -1 \}$.  Here, the future values of $c_j (\bold{p},t)$ until $t^d_j -1$ is needed to know if a different group of sufficiently many robots also stay at the task's location $l_j$ before $t^d_j$. In that case, the second team could have completed the task if it wasn't already completed, which affects the marginal contributions of the robots in the first team. 

 
 \emph{Example:} Consider the environment in Fig \ref{env} with 3 robots, all stationed at $s_1$ in cell $(2,2)$. Consider a single task $\tau_1=\{c^*_1=2,l_1=(2,2),t^a_1=0,t^d_1=4, v_1=1\}$. Let $T=4$ and let the trajectories of the robots be: 
 
 $p_1=\{(2,2), (2,2), (2,2), (2,2), (2,2)\}$,  \newline $p_2=\{(2,2), (2,2), (2,1), (2,1), (2,2)\}$, \newline $p_3=\{(2,2), (2,3), (2,3), (2,2), (2,2)\}$.
 
 In that case, the task is completed by robots $r_1$ and $r_2$ in the first time step. Note that the task cannot be completed without $r_1$ since there are no instants where both $r_2$ and $r_3$ stay at $(2,2)$. So $r_1$ should receive a utility of $U_1(\bold{p})=1$. On the other hand, if $r_2$ is removed from the system, $r_1$ and $r_3$ can still complete the task at the last time step. So $r_2$ should receive a utility of $U_2(\bold{p})=0$, although the task cannot be completed without $r_2$ in the first time step. Also, $U_3(\bold{p})=0$ since the task was completed without $r_3$.

\subsection{Learning Algorithm}
Since we designed $\Gamma_{\text{DTE}}$ as a game with the potential function equal to the total value of completed tasks, a learning algorithm such as log-linear learning (LLL) can be used to keep the long-run average performance arbitrarily close to the best possible value. In order to avoid the necessity of single-agent updates and the computation of the hypothetical utilities from all feasible actions when updating, we propose using the pay-off based log-linear learning (PB-LLL) algorithm presented in \cite{Marden12}. In this algorithm, each agent has a binary state $x_i(t)$ denoting whether the agent has experimented with a new action in cycle $t$ (1 if experimented, 0 otherwise). Every non-experimenting agent either experiments with a new random action at the next step with some small probability $\epsilon^m$, or keeps its current action and stays non-experimenting with probability ${1-\epsilon^m}$. Each experimenting agent settles at the action that it just tried or its previous action with probabilities determined by the received utilities (similar to the softmax function). Accordingly, the agent assigns a much higher probability to the action that yielded higher utility.
{\small
\begin{center}
\begin{tabular}{l l}
\rule[0.08cm]{8.5cm}{0.03cm}\\
\textbf{PB-LLL Algorithm (\cite{Marden12})}\\
\rule[0.08cm]{8.5cm}{0.02cm}\\
\mbox{\small $\;1:\;$}\textbf{initialization:} $\epsilon \in (0,1)$, $m>0$, $t=0$, $x_i(0)=0$,\\ \hspace{3cm}$\bold{p}_i(0)\in A_i$ arbitrary  \\
 \mbox{\small $\;2:\;$}\textbf{repeat} \\
\mbox{\small $\;3:\;$}\hspace{0.45cm}\textbf{if} $x_i(t) = 0$ \\
\mbox{\small $\;4:\;$}\hspace{0.9cm}with probability $\epsilon^m$ \\
\mbox{\small $\;5:\;$}\hspace{1.35cm} $ \bold{p}_i(t+1) $ is picked randomly from $A_i$\\
\mbox{\small $\;6:\;$}\hspace{1.35cm} $x_i(t+1) = 1$ \\
\mbox{\small $\;7:\;$}\hspace{0.9cm}with probability $1- \epsilon^m$ \\
\mbox{\small $\;8:\;$}\hspace{1.35cm} $ \bold{p}_i(t+1) = \bold{p}_i(t)$\\
\mbox{\small $\;9:\;$}\hspace{1.35cm} $x_i(t+1) = 0$ \\
\mbox{\small $\;10:\;$}\hspace{0.45cm}\textbf{if} $x_i(t) = 1$ \\
\mbox{\small $\;11:\;$}\hspace{1.35cm}  $x_i(t+1) = 0$\\
\mbox{\small $\;12:\;$}\hspace{1.35cm}  $\alpha = \dfrac{ \epsilon^{-U_i(\bold{p}(t-1))}}{\epsilon^{-U_i(\bold{p}(t-1))}+\epsilon^{-U_i(\bold{p}(t))}}$\\
\mbox{\small $\;13:\;$}\hspace{1.35cm} $ \textbf{p}_i(t+1)=\left\{\begin{array}{ll} \textbf{p}_i(t-1),&\mbox{ with prob. $\alpha$} \\ \textbf{p}_i(t),&\mbox{ with prob. $1-\alpha$ }\end{array}\right.$
\\
\mbox{\small $\;14:\;$}\hspace{0.45cm} \textbf{end if}\\
\mbox{\small $\;15:\;$} $t=t+1$\\
\mbox{\small $\;16:\;$} \textbf{end repeat}\\

\rule[0.08cm]{8.5cm}{0.02cm}\\
\end{tabular}
\end{center}
}
\begin{thm}
\label{Solved}
For a set of robots $R$ with the set of feasible trajectories as in \eqref{traj-set}, let $\Gamma_{\text{DTE}}=(R,A,U)$ be designed as per \eqref{actset} and \eqref{util}. If all robots follow the payoff-based log-linear learning (PB-LLL) algorithm with a sufficiently large value of $m$ in a repeated play of $\Gamma_{\text{DTE}}$, then 
\begin{equation}
\label{thmeq}
\lim_{\epsilon \to 0^+}\lim_{t^* \to \infty} \frac{1}{t^*+1} \sum_{t=0}^{t^*}f(\bold{p}(t)) = \max_{\bold{q}\in \bold{P}} f(\bold{q})
\end{equation}

\end{thm}
\begin{pf}
Since $\Gamma_{\text{DTE}}=(R,A,U)$ is a potential game with the potential function $f(\bold{p})$, in light of Theorem 6.1 in \cite{Marden12}, for sufficiently large values of $m$, PB-LLL induces a Markov chain with the limiting distribution $\mu_\epsilon$ over $A$ such that
\begin{equation}
\label{th1}
\lim_ {\epsilon \rightarrow 0^+} \mu^*_{\epsilon}(\bold{q}) >0 \Longleftrightarrow f(\bold{q}) = \max_{\bold{q}' \in A} f(\bold{q}').
\end{equation}
Note that the long term average time spent at any state $\bold{q} \in A$ converges to the  corresponding entry of the limiting distribution, i.e.,
\begin{equation}
\label{th2}
\lim_{t^* \to \infty} \frac{1}{t^*+1} \sum_{t=0}^{t^*}I(\bold{p}(t),\bold{q})=\mu^*_{\epsilon}(\bold{q}), \; \forall \bold{q}\in A,
\end{equation}
where $I(\bold{p}(t), \bold{q})=1$ if $\bold{p}(t)= \bold{q}$, and $I(\bold{p}(t), \bold{q})=0$ if $\bold{p}(t) \neq \bold{q}$. Accordingly, the long-run average of the potential function satisfies
\begin{equation}\label{th3}
\begin{split}
\lim_{t^* \to \infty} \frac{1}{t^*+1} \sum_{t=0}^{t^*}f(\bold{p}(t)) &=\lim_{t^* \to \infty} \frac{1}{t^*+1} \sum_{t=0}^{t^*}\sum_{\bold{q} \in A}f( \bold{q}) I(\bold{p}(t),\bold{q}) \\ 
&=\sum_{\bold{q}\in A}\mu^*_{\epsilon}(\bold{q})f(\bold{q}).
\end{split}
\end{equation}
Using \eqref{th3} and \eqref{th1}, we get
\begin{equation}
\label{th4}
\lim_{\epsilon \to 0^+}\lim_{t^* \to \infty} \frac{1}{t^*+1} \sum_{t=0}^{t^*}f(\bold{p}(t)) = \max_{\bold{q}\in \bold{A}} f(\bold{q}).
\end{equation}
Using \eqref{th4} together with \eqref{lem11}, we obtain \eqref{thmeq}.

\end{pf}

\section{Simulation Results}
\label{sims}
We consider the environment shown in Fig. 1 and present simulation results for two cases. 

\begin{figure}[htb]
\centering
\includegraphics[trim =15mm 10mm 0mm 0mm, clip,scale=0.27]{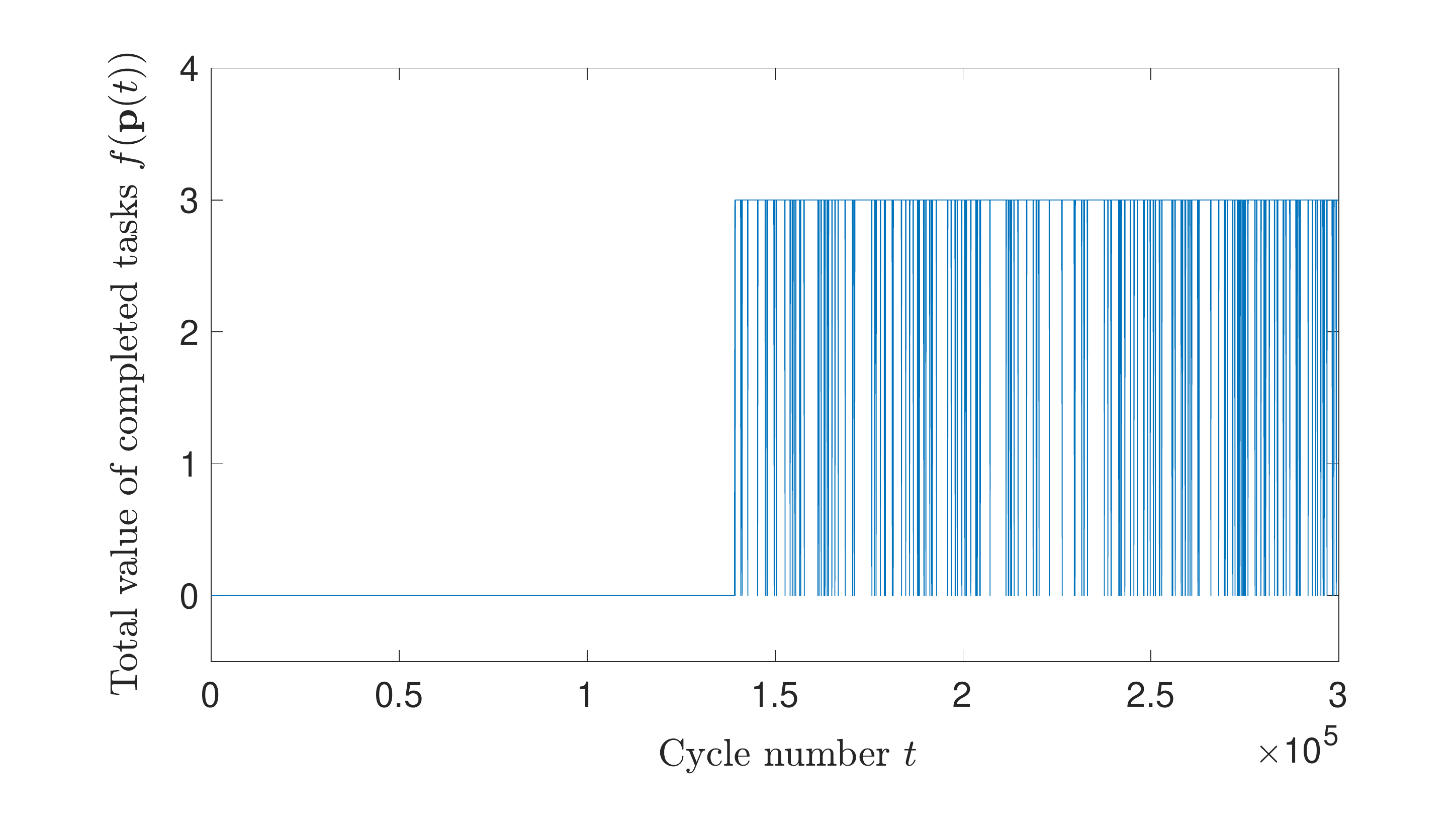}
\caption{\small Total value of completed tasks, $f(\bold{p}(t))$, in Case 1 with the proposed action sets $A_i \subseteq \bold{P_i}$ as per \eqref{actset}.  }
\label{case1a}
\end{figure}

\begin{figure}[htb]
\centering
\includegraphics[trim =15mm 9mm 0mm 0mm, clip,scale=0.25]{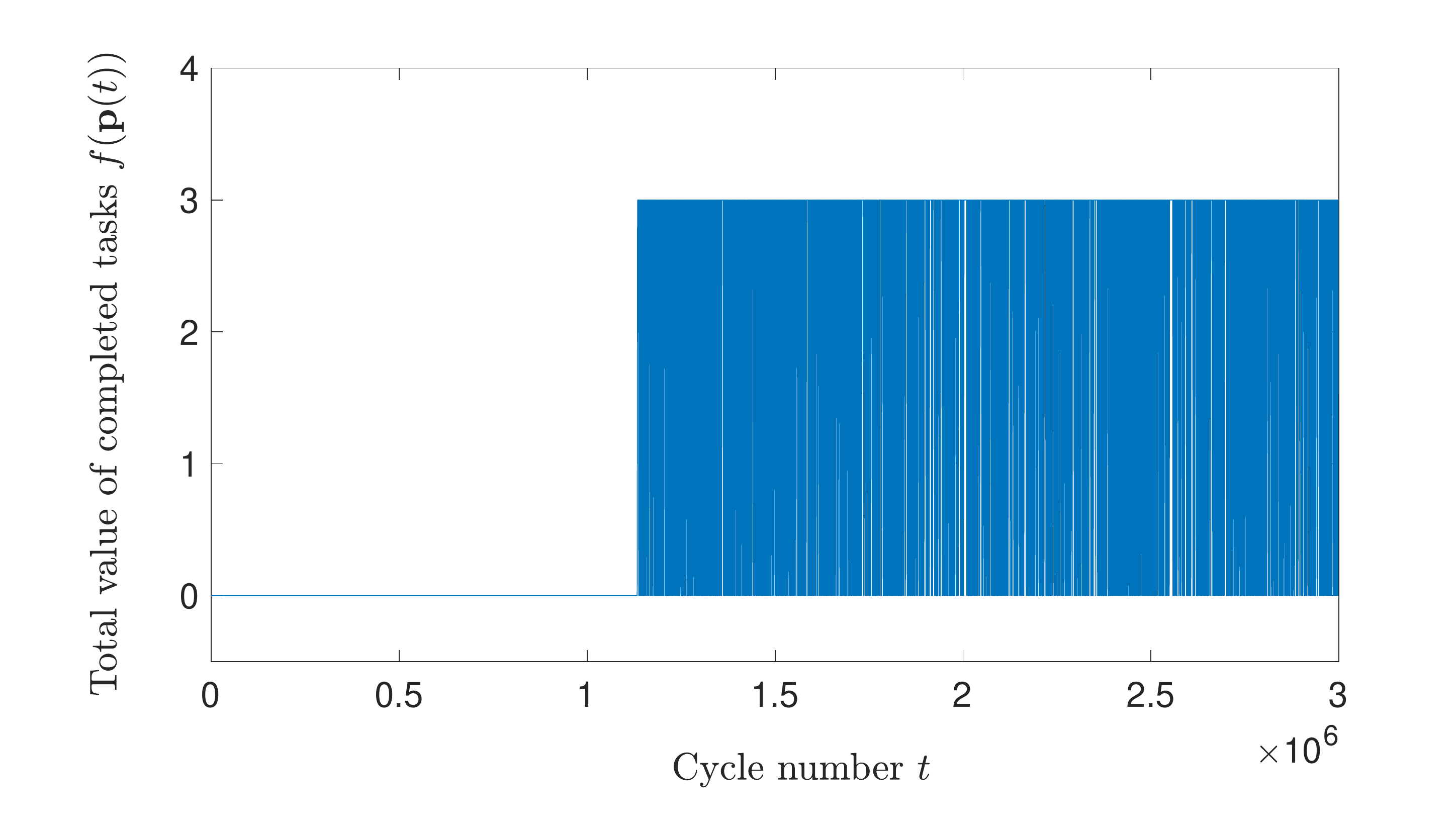}
\caption{ \small Total value of completed tasks, $f(\bold{p}(t))$, in Case 1 when the action sets are defined as $A_i$= $\bold{P_i}$. While the long-run average values are similar for the results in Figs. \ref{case1a} and \ref{case1b}, convergence to the steady state behavior is much slower when $A_i$= $\bold{P_i}$ (note the different scales on the $x$ axes of figures). }
\label{case1b}
\end{figure}

\emph{Case 1}: This case aims to demonstrate how the proposed design of action sets  $A_i\subseteq$ $\bold{P_i}$ in \eqref{actset} improves the convergence rate compared to the trivial choice of ${A_i=\bold{P_i}}$. We consider a small scenario consisting of two robots, one at station $s_2$ and $s_3$, and a single task. Each cycle has six time steps, $T = 6$. The task requires two robots to be present at cell $(6,5)$ and has the arrival and departures times as $t^a= 2$, $ t^d=5$ and a value of $v = 3$, i.e., $
    \tau_1= \{2, (6,5), 2,5,3\}$. For both simulations, robots start with randomly selected initial trajectories and follow the PB-LLL algorithm with $m=1.5$ and $\epsilon=0.007$. In this case, with the design of action sets as per \eqref{actset}, Robot 1 (station $s_3$) has $|A_1|=69$ actions, which is significantly smaller than the number of feasible trajectories $|P_1|=555$.  Robot 2 (station $s_2$) has $|A_2|=173$ actions as opposed to $|P_2|=5349$ feasible trajectories. The evolution of the total value of completed tasks (0 or 3 in this case) over cycles is shown in Fig. \ref{case1a} and Fig. \ref{case1b}. In both simulations, once the robots reach a configuration that completes the task, they maintain that most of the time. However, the scenario with the proposed $A_i$ in \eqref{actset} reaches that behavior about nine times faster than the scenario with $A_i=\bold{P}_i$. In general, this ratio of convergence rates depends on the size of the problem and may get much bigger in cases with longer cycle lengths and larger teams of robots. The repeated drops of $f(\bold{p}(t))=0$ to zero occur when robots experiment with other trajectories that result in the incompletion of the task. Due to the resulting drop in their own utility, robots almost never choose to stay at such failing configurations. For the results in Fig. \ref{case1a}, $f(\bold{p}(t))=3$ in $99.89 \%$ of the time for $t\geq 140,000$. For the results in Fig. \ref{case1b}, $f(\bold{p}(t))=3$ in $99.89 \%$ of the time for $t\geq 1,200,000$.



\begin{figure}[htb]
\centering
\includegraphics[trim =15mm 10mm 0mm 0mm, clip,scale=0.25
]{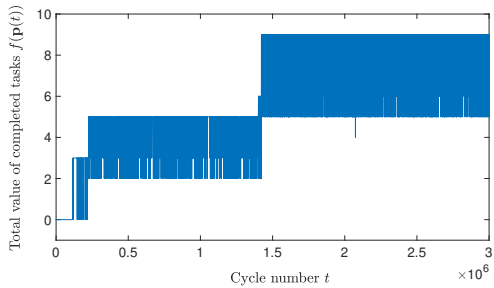}
\caption{\small Total value of completed tasks, $f(\bold{p}(t))$, in Case 2. }
\label{case2}
\end{figure}

\emph{Case 2}: We simulate a larger scenario in the same environment with more robots and tasks.  Each cycle has six time steps, $T = 6$. We define three tasks: $\tau_1$ in the previous case and two more tasks 
 $\tau_2= \{2, (4,1), 1,4,2\}$,    $\tau_3= \{3, (3,4), 1,5,4\}$. There are seven robots: two at station $s_1$, two at station $s_2$, and three  at station $s_3$. Their action sets are defined as per \eqref{actset}. Robots  start with random initial trajectories and follow the PB-LLL algorithm with $m=1.8$ and $\epsilon=0.007$.
Fig. \ref{case2} illustrates the evolution of the total value of completed tasks over cycles. We observe that the robots successfully complete all three tasks in $ 99.91\% $ of the time after ${t=1,500,000}$.

\section{Conclusion}
\label{conc}
We presented a game-theoretic approach to distributed planning of robot trajectories for optimal execution of cooperative tasks with time windows. We considered a setting where each task has a value, and it is completed if sufficiently many robots simultaneously spent one unit of time at the necessary location within the specified time window. Tasks keep arriving periodically over cycles with the same specifications, which are unknown a priori. In consideration of the recharging and maintenance requirements, the robots are required to start and end each cycle at their assigned stations and they try to maximize the value of completed tasks by planning their own trajectories in a distributed manner based on their observations in the previous cycles. We formulated this problem as a potential game and presented how a payoff-based learning algorithm can be used to maximize the long-run average (over cycles) of the total value of completed tasks. Performance of the proposed approach was also demonstrated via simulations. 

\bibliography{MyReferences}

\end{document}